Consciousness is Pattern-Recognition: A Proof
Copyright Ray Van De Walker 2016, Licensed under Creative Commons License
Attribution 4.0 International License, as specified at
http://creativecommons.org/licenses/by/4.0/legalcode
Author: rgvandewalker –at- yahoo –dot- com
orcid:0000-0001-9072-7390

Abstract:
This is a proof of the strong AI hypothesis, i.e. that machines can be conscious. It is a phenomenological proof that pattern-recognition and subjective consciousness are the same activity in different terms. Therefore, it proves that essential subjective processes of consciousness are computable, and identifies significant traits and requirements of a conscious system. Since Husserl, many philosophers have accepted that consciousness consists of memories of logical connections between an ego and external objects. These connections are called "intentions." Pattern recognition systems are achievable technical artifacts. The proof links this respected introspective philosophical theory of consciousness with technical art. The proof therefore endorses the strong AI hypothesis and may therefore also enable a theoretically-grounded form of artificial intelligence called a "synthetic intentionality," able to synthesize, generalize, select and repeat intentions. If the pattern recognition is reflexive, able to operate on the intentions, and flexible, with several methods of synthesizing intentions an SI may be a particularly strong form of AI. Similarities and possible applications to several AI paradigms are discussed. The article then addresses some problems: The proof's limitations, reflexive cognition, Searles' Chinese room, and how an SI could "understand" "meanings" and "be creative."

This paper directly proves the "strong AI hypothesis" that consciousness is computable. Also, this proof describes critical features of the algorithms of consciousness, which may help practical AI development and testing.

One problem with any such proof is that conventional tests of consciousness are subjective, thus the proof must be at least half phenomenal. The required phenomenal analysis seems to have stymied many researchers.

I'd like to describe the proof that persuaded me. I haven't seen it anywhere else, so as far as I know, it is original.

Briefly, a philosophically respectable position is that consciousness is always consciousness… *Of. Some. Thing.* There is a substantial body of philosophy, Phenomenology, which studies the connection between a perceiver, and the object, i.e. the meaning of that critical little word "of." Phenomenology is often defined as the study of experience.

Some evidence that phenomenology may be relevant to AI is that by 1930, phenomenologists had uncovered the complexity of natural human intelligence. They recoiled in horror at the "vast field of toilsome discoveries" of which logic, mathematics and epistemology were small *parts*.[1] This is clearly parallel with more-modern experiences in practical AI.

Edmund Husserl, who cast phenomenology in its modern terms, describes a consciousness as a memory or stream of experience of the logical connections (or "intentions") between an "ego" and other things.[2] His proof and evidence is widely respected by philosophers, and is beyond the scope of this paper.

Intentions (connections between things and an ego) include perception, belief, observation, desire, communication and will. All of these are described as "of," "with," "about" or "to." In people, intentions seem to occur about 10 times per second. Husserl claims that consciousness consists of sequential memories of intentions.

If Husserl's proofs are right, the practice of strong AI should be phenomenological engineering: The design of consciousness is the design of intentions between a self and objects, recorded in a memory.

If the connection between ego and object is an "intention," then a *mechanism* that synthesizes intentions would be a "synthetic intentionality," or "SI."

To make a rigorous proof, a basic research tool of phenomenology is needed, an introspective mental operation called "bracketing." The name is from the idea of putting some part of one's experience into "brackets," and mentally pretending that it and its logical consequences don't exist. Bracketing is essential to the proof that follows.

In phenomenal experiments, one brackets some part of one's experience, and then observes how one's experience would be different. The really unique thing about phenomenal experiments is that they require no equipment and little preparation. So, they're actually better than logic for making a proof. With logic, one has to start from agreed premises. Phenomenal truths are objective because they're so easy for individuals to reproduce.

The utility of bracketing is that one can examine the conceptual structure of one's experiences in detail. For example, one can bracket the color or smell of an apple, and it still can be an apple. One cannot bracket "edibility" in an apple, and retain "appleness." If one does, then a moist wax model of an apple becomes phenomenally equivalent to a real apple. This shows the interesting fact that edibility is part of the mental concept of an apple.

*Briefly:*

The proof uses bracketing to analyze the "object-ground" problem. Briefly, this problem asks: "What's an 'object'?" and, "How do people separate objects from backgrounds?"

One way to investigate this is to bracket all objects. This experiment has the interesting effect that what's left is ground, mere qualia (or "sense data").

There are some further interesting side-effects. When I do this, I have to remove all thought from consideration, because thinking is precisely "about" "things." In order to think, or apparently to do anything "conscious," people have to make logical connections between things, that is, "objects," and themselves, their "ego."

It occurred to me that the process of synthesizing intentions, i.e. separating "objects" from the "ground" was precisely the problem that AI researchers call "pattern recognition."

That is, pattern recognition is consciousness. Therefore, since pattern recognition can be computed, then consciousness can also be computed. That is, the "strong AI hypothesis" is confirmed.

The identity of consciousness and pattern recognition has already been recognized by many AI researchers, but the lemma that it proves computability of consciousness has been neglected.

A more detailed phenomenal analysis yields not only a more rigorous proof, but also identifies essential features of consciousness and its necessary algorithms.

*Here's the details:*

For an example to generalize from, let's imagine a very simple pattern recognition program. Let's say that it finds square-shaped patches of zeros in a square array of numbers. I am sure that this is within the state of the art, because I could program it myself. This might even be useful, if the array of numbers was from a video camera, or had mathematical interest.

Now, for lemma A, let's bracket each piece of consciousness as it is found in the pattern recognition program.

*Lemma A1:*

If one brackets the program's concept of "*objects*" a *square* can't be recognized, because it's an object. The logical consequences, i.e. the variables identifying it, must be removed from consideration, and therefore from usage. The bracketed program cannot perform its function.

*Lemma A2:*

If one brackets the program's *connection* between the concept of square, and a position in the array, the logical consequences, i.e. the variables owned by the program that identify or locate the square, must be removed from consideration, and therefore from usage. The bracketed program cannot perform its function.

*Lemma A3:*

If one brackets the concept of a "recognizer" (i.e. an ego) from the program, then the purpose and meanings of the program's outputs are lost, and therefore the program can't perform its function. It might still produce data, something like "qualia," perhaps, but never information. (Note that qualia as such lack memory, ego and a logical connection, and are not enough to produce consciousness.)

*Lemma A4:*

If one brackets the program's *memory* of such a connection, the justification for any belief is not available. The program may produce data, but there is no evidence from it. In particular, there's no way to decide that some sense-data is or was a square. Again, the progam can't perform its function.

*Lemma A, Evidence:*

I think it's clear that almost all pattern recognition programs would have similar issues, if the parts of consciousness were bracketed in similar very general ways. There might be pathological cases that don't reduce, but they will be remarkably interesting in their own right for their very peculiar properties. These special cases might be ways to produce exceptionally stable synthetic intentions, or especially low-cost or well-performing implementations.

For the general case, Lemma A shows that bracketing significant parts of consciousness in a pattern recognition program causes the pattern recognition program to fail to recognize. It is no longer a "pattern recognition" program. These parts are essential, that is, required by definition.

*Lemma A, Conclusion:*

Thus, by eliminating the concepts which are essential to consciousness: Any of: objects, the connection, the former of the connection or the memory of the connection, one eliminates the *equally essential parts* of pattern recognition.

*Lemma B, Evidence:*

Now, for lemma B, let's bracket pattern recognition from consciousness.

Imagine, a human being, *someone* who is indisputably conscious, such as yourself. Bracket your pattern recognition. That is, pretend to yourself that "everything which was logically dependent on pattern recognition" ceased to exist.

*Lemma B1:*

One will discover that one cannot recognize any objects in such a state; Consciousness is removed from consideration because it forms intentions with (logical connections to) objects.

*Lemma B2:*
Further, the intentions, the logical connections are gone as well. There are no recognized objects to which they can attach.

Tellingly, even Husserl's "transcendentally pure consciousness" (when one's consciousness is conscious only of itself) is removed from consideration, because one must *recognize* one's own consciousness as being different from the other items of one's mental landscape.

*Lemma B3:*
In this state, there may be sense-data, so-called "qualia," but there is no narrative, even as a sequence of connected mental pictures. In a real sense, formation of an observation is impossible, and therefore there is no *observer (i.e. no ego)*. Arguably, the consciousness itself does not exist in this state. That is, there is no consciousness, in a different way. (Qualia as such are not sufficient to identify consciousness.)

*Lemma B4:*
Memory becomes impossible, because recognition experiences objects in time and space. When recognition is removed from consideration, space, sequence and time are also removed from consideration. Memories depend on these, and are also removed from consideration.

Why do I consider time and space essential items for memory? Well, a simple example is food. If you remember food, the memory is utterly useless unless you can use the memory to get the food. Arguably, such a mental phenomenon without time and location is so useless that it's not a memory.

This is rather a weak spot. Often people substitute a discovery procedure for a reliable memory. We look for restaurants on the street or net, or use a cook-book or phone-book.

However, I would argue then that what we are responding to is not a memory, but a hope, and we're trying to convert the hope into a plan. This may eventually turn into a memory, but it simply isn't one *yet*.

*Lemma B Conclusion:*
All the items of consciousness are removed from consideration when pattern recognition is bracketed.

*Main lemma, combining A & B:*
The logic:
If not A implies not B and not B implies not A, then B implies A and A implies B.
That is, A and B are biconditionally equivalent.
Restated: The items A and B have the same logical effects in different terms.

By this logic, pattern recognition in objective technical terms has the same effects as consciousness, in different, subjective terms. Therefore, since many forms of pattern recognition are computable, then parallel forms of subjective consciousness are also computable and vice-versa.

An even more detailed phenomenal examination may yield more insight about how to implement more human-like AIs. But, there's enough to move forward. Also, useful SIs might not need to resemble human cognition much.

So, let's speculate about how to apply these logical identities.

A memorized sequence of intentions, what we subjectively call an "experience," might be selectively replayed by a synthetic intentionality as it uses its pattern recognition to unify a sequential graph of intentions with its sense-data.

With simple feedback and some reflexive processes, a synthetic intentionality might select its sequence of intentions to search for and repeat subjective experiences. That is, given feedback equivalent with the qualia of pain and pleasure, and a method to find and replay intentions, a synthetic intentionality can have an experience that is subjectively equivalent to learning.

A fruitful application of an automaton's pattern recognition might be to reflexively apply pattern recognition to its own memories and internal operations. In this way it could even learn to improve its general problem-solving methods.

An SI requires only one data structure, a graph of "intentions." Intentions can be passive or active, because the type of connection (between ego and object) of an intention can vary. Also the data structure of intentions is finite, closed, defined by the hardware (input, output and reflexive) that the SI is designed to operate. Given full reflexive access, an SI might even be able to compile intentions into optimized code for its CPU or other computational substrate.

That is, in subjective terms, continuing effort by a programmer is not needed for a synthetic intentionality to learn by experience and improve itself.

When intentionally-guided pattern recognition failed, a synthetic intentionality could fall back to evolutionary searches for intentional sequences,[5] Bayesian-guided searches of a stochastic space of intentional sequences, or, when lacking data, even random generation and recording of intentions. The result could be a very strong form of AI, whose intentions, intentional sequences and later algorithms are not limited by its starting algorithms.

These might be intentional, literally conscious improvements. However, if unconscious methods were used (e.g. evolution of intentional graphs) these unconscious reflexive processes might occur in a subjective experience like sleep, to avoid interference with real senses and effectors during the reorganization of the intentions.

We'd expect successes in AI systems that resemble synthetic intentionalities (SIs), and failures as they depart from that model. The canonical form of a synthetic intentionality might synthesize and store a database describing a graph of intentions, then apply pattern recognition to realia to select a graph or intentions to predict the future in a limited way and cope with the future. This canonical form might be useful in some applications: Explicit reasoning such as mathematical proofs, heuristic descriptions, knowledge transfer, etc.

A synthetic intentionality thus resembles a frame-based[4] AI, except that an SI fixes some of frames' practical issues by using explicit pattern recognition to automatically provide new frames, linkage between frames and data, and other context.

An SI is amenable to genetic programming[5] with the advantage that an SI's proven-sufficient, defined-by-I/O data structure removes any necessity to manually design new data structures.

A compact, fast, parallel implementation of an SI might be neuromorphic. It could be like Hawkins' hierarchal temporal memory[6] (an HTM has hierarchies of stochastic forward-predicting state-machines.) As in an HTM, a hierarchy can multiply a (relatively) small number of intentional graphs (thousands) into many distinct intentions, economically yielding trainable, flexible, adaptive behavior. However, an SI's reflexive consciousness (i.e. "imagination" or "abstraction") could speed learning compared to the unconscious training of current HTMs. Also, the canonical form of SIs may permit direct design of HTMs, given some translation from canonical to HTM.

Conversion between an easily analyzable database of intentions and a neuromorphic implementation might be by something like Tononi's integrative process(es), that produce transition probability matrices.[8] This would permit a-somatic design or training of SIs.

A reverse elaborative process would decompile neural or neuromorphic data into a graph of intentions, permitting transfer, design, functional composition, optimization and synthesis of SIs.

This proof and its schemes have some issues. First, the proof has reasonably clear limitations in its fidelity because of its incompleteness and unrealistic simplicity.

As to completeness, the proof may fail to describe many parts of human consciousness. But, many major, subjective tests of consciousness will be satisfied, because the proof leverages decades of research in phenomenology. Also, it's not reasonable to expect humans and machines to have *identical* consciousnesses without targeted research and development.

The actual complexity of subjective phenomena and implementations may prevent high fidelity in basic implementations. However, R&D can improve a practical implementation's fidelity until it's valuable in practice.

A theoretical problem is to define "pattern recognition" well enough to implement an SI. Ideally, such a definition would be mathematically complete, closed over all possible experiences of the SI. However, since the SI is computable, this seems to require that the SI's mental system be both consistent and self-proving, which Gödel proved is likely impossible.

Luckily, we have examples of pattern recognition. Using these, phenomenological engineers can build SIs without a general definition. The story here would be something like, "Smarter SIs will recognize more items and types of items, and therefore smarter SIs will have expanded forms of consciousness." This puts SI design into a continuum of technique by which SIs can be improved like any other technical artifact. Eventually a limited theory of pattern recognition may be possible, and bring many improvements.

But, can an SI possibly be conscious? Many philosophers argue that computers are "syntactic," that is, they perform only symbol manipulation. Then these philosophers prove that consciousness is nonsymbolic, and therefore "can't be performed by computers." This is Searle's "Chinese room."[9]

Searle is qualitatively right. Pattern recognition, therefore consciousness, is generally an analog process. That is, its inputs are smoothly varying quantities from sensors that interact with some nonsymbolic "real world."

However, there's no profound problem in turning those quantities into streams of numbers and processing the numbers. Electrical engineers frequently use "digital signal processing" in place of analog circuitry. It works well. Almost all digitally-recorded music "sounds like music." Even when it goes wrong, it sounds like "badly recorded" music. There are technical deficiencies in digitization, but they are well-understood sources of error, characterized mathematically using the "Z transform" to manage "sampling error," "quantization error" and "frequency response."

The "non-syntactic" nature of consciousness thus seems amenable to normal engineering tradeoffs between the costs and convenience of digital and analog designs. Just design the desired pattern recognition algorithm. Then use the cheaper of analog or digital implementation.

Reflexivity is also an issue. It's common to believe that only a conscious being can perceive meaning. So, "recognizing concepts" is at the core of consciousness. Some philosophers still argue about whether one can recognize a concept. But, the proof says that consciousness and pattern recognition have identical consequents, so if one can be "conscious of" a concept, one can "recognize" it. So, the existence of the philosophers' argument *itself* indicates that people can be conscious of a concept, and therefore can recognize it.

There's also real dispute among philosophers about what "meaning" is, and therefore whether automata can perceive it. Let's use Wittgenstein's assertion that meaning is how one uses words.[10] This explains why meaning requires exactly a set of words and an interpreter. Still, Wittgenstein leaves open what one is doing, and what words are.

As programmers might note, Wittgenstein's definition is shockingly close, perhaps even practically close, to such computer-science subjects as Turing-equivalent instruction sets and syntax-directed compilation. But, Wittgenstein's definition is too vague to be inherently mechanistic. Note that the interpreter could even be a human soul increasing in spiritual beauty because of the nourishment of God's word.

Compiling to intentions can resolve these issues. An SI could consciously understand meaning by synthesizing new intentions from a stream of words or other symbols, and linking it to its preexisting set of intentions. And, in a crucial test of consciousness, an SI can *misunderstand*, by mislinking intentions to its preexisting intentions, and *correct* its misunderstanding by relinking them more accurately.

Wittgenstein's words could be absolutely anything that is not already in the interpreter, including the experience of eating an apple. The *memory* of such an experience could be in the interpreter, but the *experience* cannot be, because it occurs only when the apple goes into the interpreter's mouth. (Note that experience requires particular *equipment*: e.g. a "mouth." This is evidence that the physical structure of the interpreter is crucial to symbolic interpretations of experience.) Synthesizing intentions can solve these issues, too.

It's also common to believe that creativity is an essential element of consciousness. How might an SI be creative? From my personal observations, creative people do something very like rolling weighted mental dice when making contingent decisions.

An SI can handle such contingent decisions by using what amounts to an infinitely-divisible roulette wheel, in which different contingent choices of paths through a graph of intentions are assigned to each slice of the wheel. And, there can always be some probability of a totally random choice.

When the SI starts, totally random choices might be a large part of each intention. But, if a result is sufficiently unpleasant, that choice can be avoided in the future by reducing that choice's slice of the probabilities. The contingent choices without unpleasant consequences will remain contingent, making the SI creative throughout its behavior.

Pattern recognition also appears to be the foundation of *all* symbol manipulation. The way that people perform symbolic manipulation is that they recognize patterns of symbols, and transform them into other patterns.

Even digital computers use analog pattern recognition, because every binary switch has to recognize whether its control is *on* or *off*. If this is right, then computers are *already* conscious, in trivial ways. The same sense of consciousness over time is in every thermostat: Too hot or too cold, with some memory of other states, and something to do about them.

So, synthetic intention is not only possible, *it already exists.* Doing it more skillfully, and linking it to languages, space, time and other qualia ("sense data") seems perfectly possible.

Citations